# Hetero-Center Loss for Cross-Modality Person Re-Identification

Yuanxin Zhu[a], Zhao Yang[a,*], Li Wang[a], Sai Zhao[a], Xiao Hu[a], Dapeng Tao[b]

[a]School of Mechanical and Electric Engineering, Guangzhou University, Guangzhou, China
[b]School of Information Science and Engineering, Yunnan University, Kunming, China

*Abstract* –Cross-modality person re-identification is a challenging problem which retrieves a given pedestrian image in RGB modality among all the gallery images in infrared modality. The task can address the limitation of RGB-based person Re-ID in dark environments. Existing researches mainly focus on enlarging inter-class differences of feature to solve the problem. However, few studies investigate improving intra-class cross-modality similarity, which is important for this issue. In this paper, we propose a novel loss function, called Hetero-Center loss (HC loss) to reduce the intra-class cross-modality variations. Specifically, HC loss can supervise the network learning the cross-modality invariant information by constraining the intra-class center distance between two heterogenous modalities. With the joint supervision of Cross-Entropy (CE) loss and HC loss, the network is trained to achieve two vital objectives, inter-class discrepancy and intra-class cross-modality similarity as much as possible. Besides, we propose a simple and high-performance network architecture to learn local feature representations for cross-modality person re-identification, which can be a baseline for future research. Extensive experiments indicate the effectiveness of the proposed methods, which outperform state-of-the-art methods by a wide margin.

## I. INTRODUCTION

Person re-identification is an image retrieval problem, aiming to match pedestrian images across multi-cameras views [1]. Most of the existing works [2] [3] [4] [5] [6] [7] focus on matching RGB images. However, there are some limitations for the RGB based images re-identification task. For example, criminals often gather information in the day and execute crimes in the night. Fortunately, most the recent surveillance cameras can capture infrared images at night, which can provide valid information for some related tasks. In this case, the traditional method can not address this kind of problem properly, because there is a huge gap between infrared (IR) images and RGB images, as shown in Figure 1. Comparing to RGB images, IR images lose rich color information, which is important in RGB-based person Re-ID methods. In addition, the spectrum between IR and RGB images is different. So, the method for RGB-based person Re-ID can not be adopted in RGB-IR cross-modality person Re-ID problem effectively [8].

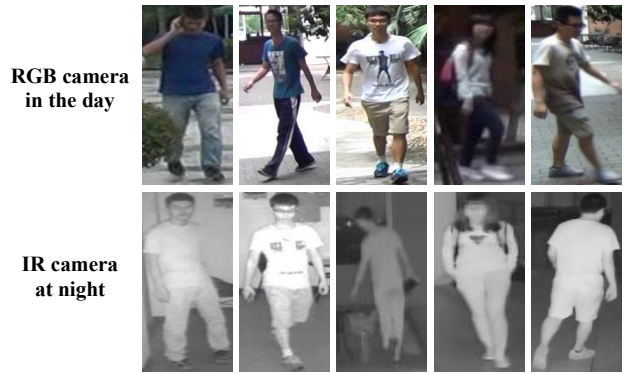

Figure 1. Examples of RGB images and infrared images in SYSU-MM01 RGB-IR [8] dataset.

To address this issue, some methods have been proposed in this field. Wu et al. [8] released a large-scale cross-modality person Re-ID dataset and proposed a deep one-stream architecture named zero-padding network. In the training stage of zero-padding network, Cross-Entropy (CE) loss function is used to supervise the network. Then Ye et al. [9] proposed a two-stream network architecture called TONE, in which CE loss and contrastive loss are used for training. As a ranking loss, contrastive loss is complementary to CE loss. Then, Ye et al. [11] used triplet loss instead of contrastive loss to train an improved two-steam model named BDTR based on TONE, because contrastive loss is of weak flexibility in the feature embedding learning. Contemporarily, Dai et al. [10] also adopted the joint supervision of triplet loss and CE loss to train a generative adversarial network named cmGAN which can learn modality-invariant feature representation.

However, most of the above-mentioned methods focus on enlarging inter-class discrepancy of features and ignore improving the intra-class cross-modality similarity. The two objectives are equally important for this issue. In this paper, we design a novel loss function specifically for the problem, called Hetero-Center (HC) loss, which constrains the intra-class center distance between two heterogenous modalities. The loss function can force the network extracting the invariant modality-shared information rather than inconstant modality-specific information from heterogeneous images to form the feature descriptors. To achieve the two aims simultaneously, we adopt the joint supervision of the HC loss and CE loss to train the network. Both of them can be minimized by standard optimize algorithms, e.g. Stochastic Gradient Descent (SGD) [13].

*Corresponding author
Email address:yangdxng100@126.com (Zhao Yang)

Besides, we propose a network framework called Two-Stream Local Feature Network (TSLFN) which learns local feature representations to solve the problem. The architecture is divided into two individual branches to extract features in two modalities. Each branch contains a backbone network, which outputs a feature map with rich image information. Then, conventional average pooling layers are employed to uniformly split the feature maps into several stripes for local feature extraction. To project the features from different modalities into the same subspace, we use a share-weight fully-connected layer for corresponding stripes in two branches. The experiments demonstrate that TSLFN with HC loss achieves state-of-the-art performance in this field, which far exceeds other methods.

The main contributions of this paper can be summarized as follows:
1. We design a novel loss function (named HC loss) to constrain the distance between two centers of heterogenous modality. HC loss forces the network improving the intra-class cross-modality similarity. With the joint supervision of HC loss and CE loss, the network extracts modality-shared information to form discriminative feature descriptors.
2. We present a network structure to learn local feature representation. To the best of our knowledge, it is the first attempt to learn local feature representations in the field of cross-modality person Re-ID. Due to its simple and effective architecture, the network can be a strong baseline for future research.

## II. RELATED WORK

In the field of person Re-ID, most of the works focus on dealing with the matching problem in RGB domains. Those methods could be divided into three categories: hand-craft feature representation [3] [6] [14] [15] [16] [17] [23], distance metric learning [18] [19] [20] [21] [22] and deep learning [24] [25] [26]. A detailed literature review for RGB-based person Re-ID can be found in [1]. However, the performance of the above methods on RGB-IR cross-modality person Re-ID problem is poor, because there is a large gap between RGB domains and IR domains. To deal with the cross-modality retrieval problem, the following approaches were proposed.

Wu et al. [8] firstly defined the problem of cross-modality person Re-ID, and released a large-scale cross-modality person Re-ID dataset, named SYSU-MM01. To address the problem, they discussed the difference between three commonly used cross-domain models: asymmetric FC layer, one-stream, and two-stream network structures. Based on the discussions, they proposed an improved one-stream network architecture named zero-padding network, which converted images from RGB color space to gray color space in the preprocessing phase. Then, a gray image was placed in the first channel and a zero-padding image was placed in the second channel. By contrast, an infrared image was placed in the second channel and a zero-padding image was placed in the first channel. The purpose of zero-padding network was to increase domain-specific nodes in the network, which provided extra flexibility for the network.

Ye et al [9] [11] pointed out that cross-modality person Re-ID suffered from cross-modality and intra-modality variations simultaneously. Based on the point of view, Ye et al. [9] proposed a hierarchical metric learning method called HCML for cross-modality matching. The objective of HCML was to learn a kernel matrix. By the matrix, features were projected into a subspace, in which the two variations were minimized as much as possible. To learn the matrix, the formula of HCML contained two optimization terms, which were modality-specific metric term and modality-shared metric term. The aim of the first term was to constrain the features extracted from the same modality as compact as possible, which could reduce the intra-modality variations. For the second term, the aim was to improve the discriminative power of the features extracted from two modalities for pedestrian identity.

Besides, Ye et al. [9] proposed a two-stream convolution network structure named TONE. In the training stage of TONE, the joint supervision of contrastive loss and CE loss was adopted to train TONE. Based on TONE framework, Ye et al. [11] proposed an improved two-stream network structure named BDTR. The difference between the two networks was that BDTR used triplet loss [27] to supervise the training of network instead of contrastive loss. Since contrastive loss used a fixed margin for all negative images, which was quite restrictive for the feature distribution, damaging its robustness for noisy samples in feature embedding learning [43]. Comparing with contrastive loss, triplet loss only forced negative images to be farther away than positive images, which was more robust for distortions.

Dai et al [10] proposed a novel method termed as cmGAN, which achieved the advanced performances in the field. The method was based on generative adversarial network (GAN) [28] [40], which consisted of a generator and a discriminator. In cmGAN, generator extracted features from two modalities, which were input into the discriminator. The aim of the discriminator was to distinguish whether the input features were from RGB modality or infrared modality. In contrast, the aim of the generator was to extract features which could not be correctly judged by the discriminator. By training the two networks with opposite aims, cmGAN could learn modality-invariant feature representations. In the training phase, cmGAN adopted the joint supervision of triplet loss and CE loss as [11].

## III. METHODS

### A. Problem description

In heterogenous images, the appearance of a person consists of modality-shared information (e.g. contours,

textures) and modality-specific information such as colors. The former information is the invariant information existing in two modalities, which should be extracted by the network to form the feature descriptor, due to its robustness for modality changes. The latter information only exists in specific modality or is inconstant with modality changes, which reduces the feature similarity between two heterogenous samples of the same identity.

Since cross-modality person re-identification is a verification problem, we compute the similarity of features extracted by the network to match the pedestrian images between two modalities. Hence, the aims of the network in the training procedure are to enlarge the inter-class discrepancy and to improve the intra-class cross-modality similarity. So, the features should contain modality-shared information as much as possible to bridge the gap between two modalities, which improves the intra-class cross-modality similarity. However, traditional loss functions can not supervise the network to extract modality-shared information. For instance, CE loss function is computed as

$$L_S = -\sum_{i=1}^{K} \log \frac{e^{W_{y_i}^T x_i + b_{y_i}}}{\sum_{j=1}^{n} e^{W_j^T x_i + b_j}}, \quad (1)$$

where $K$ denotes the batch size, $x_i$ denotes the features extracted by $i^{th}$ sample belonging to the $y_i$ class, $W_j$ denotes the $j^{th}$ column of the weights, and $b$ is the bias term. From the definition of Cross-Entropy, we can observe that the objective of Cross-Entropy is to extract identity-specific information for classification. However, the loss function does not constrains the network to extract modality-shared information effectively to form the feature descriptor, because some modality-specific information is also the identity-specific information conducting the network to correctly predict identity. For example, clothes color attribution is a strong signal to predict the true label, which is probably extracted by the network with the supervision of CE loss to form the descriptors. However, the color attribution is inconstant with modality changes, and the operation of extracting the information to form the descriptors is contradictory with the aim of improving intra-class cross-modality similarity. Therefore, CE loss can not achieve the vital objective, intra-class cross-modality as much as possible. Analogously, most of the conventional loss functions can not meet the request of cross-modality person Re-ID.

To intuitively demonstrate the disadvantage of conventional loss functions, in Figure 2 we show typical feature distributions with the supervision of CE loss. From the figure, we observe the phenomenon that the features of different classes are separated correctly. However, the feature distributions of different modalities exist a huge gap in each class, which is reflected from the considerable center distance between two modalities in the figure.

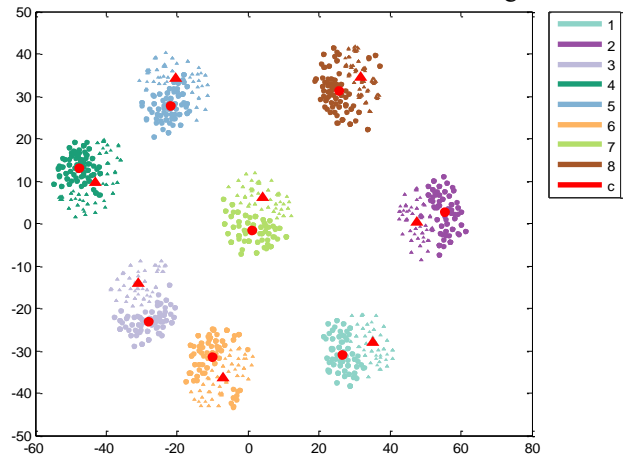

Figure 2. The distribution of features extracted by the baseline model (its architecture is the baseline model which will be mentioned in III section) only with CE loss. The feature is from 770 RGB images and 300 infrared images of 8 identities in SYSU-MM01, whose dimension of features is reduced to 2 by t-SNE [42]. Points with different colors denote features belonging to different identities. Points of different shapes denote features extracted from images of different modalities. The red points with different shapes denote the feature centers of different modalities in each identity.

*B. Hetero-Center Loss*

In this subsection, we propose our loss function to improve the intra-class cross-modality similarity. Intuitively, we want to constrain the distance between two modality feature distributions in each class. However, it is hard to compute the distance between two feature distributions, so we penalize the center distance between two modality distribution instead of the distance between two modality distribution. To this end, we propose Hetero-Center (HC) loss as formulated in the following equation

$$L_{HC} = \sum_{i=1}^{U} \left[ \left\| c_{i,1} - c_{i,2} \right\|_2^2 \right], \quad (2)$$

where $c_{i,1} = \frac{1}{M} \sum_{j=1}^{M} x_{i,1,j}, c_{i,2} = \frac{1}{N} \sum_{j=1}^{N} x_{i,2,j}$ denotes the centers of feature distribution of RGB modality and infrared modality in the $i^{th}$ class. $U$ denotes the number of classes, $M$ and $N$ are the numbers of RGB images and infrared images in the $i^{th}$ class. $x_{i,1,j}$ and $x_{i,2,j}$ denotes the $j^{th}$ RGB image and infrared image in the $i^{th}$ class. Ideally, the centers of two modalities in every class are supposed to be updated when the weights of the model are updated in each epoch. In this case, we need to consider every sample to learn the two centers of each class in each iteration, which requests massive and unpractical computational cost.

To solve the problem, we conduct two efficient modifications inspired by [29] [12]. First, we compute two modality centers of each class in a mini-batch rather than in the total training set. Consequently, the constraint on center distance comes into force in the mini-batch, instead of the entire training set. Second, to make the constraints equivalent in different ranges, we propose an improved mini-batch sampling strategy based on Ye et al [11]. In each iteration, we randomly choose $L$ identity from the training set. Then, we randomly select $T$ RGB images and $T$ infrared images of each chosen identity to form a mini-batch, so its size is $2 \times L \times T = K$. In this way, the modality centers in a mini-batch are computed from multiple features and the sample size of each class is the same, which is important to avoid the perturbations caused by class imbalance. And, due to the random sampling in multiple iterations, the local constraint in the mini-batch has the same effect as the global constraint in the entire training set.

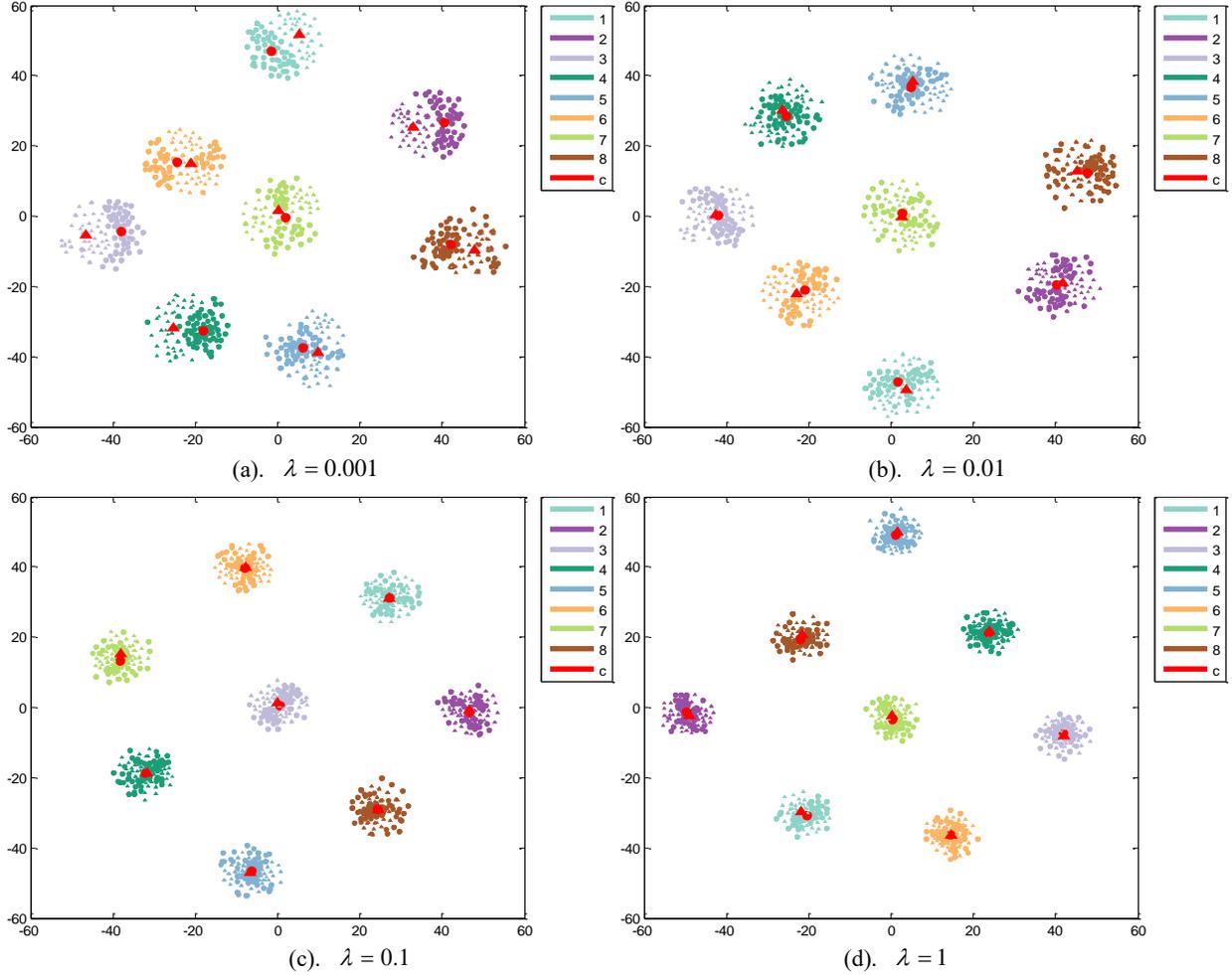

Figure 3. The feature distribution extracted by the baseline model with the joint supervision of CE loss and HC loss. The points with different colors and different shapes denote features of different modality belonging to different identities. The red points denote the feature centers of different modalities in each identity. Different $\lambda$ leads to different feature distributions. We can observe that with the increase of $\lambda$, the feature distributions of different modality are pulled closer and the distance between two feature centers of different modalities is smaller.

Since HC loss only constrains the center distance in each class to improve the intra-class cross-modality similarity, it can not supervise the network learning discriminative feature representation to enlarge the inter-class discrepancy. By considering the two key objectives for cross-modality person Re-ID, we adopt the joint supervision of HC loss and CE loss. The overall loss function is given as

$$L = L_S + \lambda L_{HC}$$
$$= -\sum_{i=1}^{K} \log \frac{e^{W_{y_i}^T x_i + b_{y_i}}}{\sum_{j=1}^{n} e^{W_j^T x_i + b_j}} + \lambda \sum_{i=1}^{U} \left[ \left\| c_{i,1} - c_{i,2} \right\|_2^2 \right], \quad (3)$$

where $\lambda$ is a hyperparameter for balancing the two loss functions, which is regarded as the weight of HC loss in the overall loss. Figure 3 shows the feature distributions with different $\lambda$, from which we can intuitively observe the influence of HC loss in the course of training. With the increase of $\lambda$, the feature distributions of different modality are pulled closer and the distance between two feature centers of different modalities is smaller, which means that the learned feature representations are more consistent for different heterogenous images and the network is more inclined to extract modality-shared information to form the feature representations. The trend demonstrates that the intra-class cross-modality similarity is increased with the supervision of HC loss.

The overall loss can be optimized with standard optimization algorithms (e.g. SGD), because the gradients of $L_{HC}$ with respect to $x_i$ can be directly solved as

$$\frac{\partial L_{HC}}{\partial x_{i,1,j}} = \frac{\partial L_{HC}}{\partial c_{i,1}} \frac{\partial c_{i,1}}{\partial x_{i,1,j}} = \frac{2}{N}(c_{i,1} - c_{i,2}). \quad (4)$$

By the same principle, the gradient of $x_{i,2,j}$ can be also computed. When the models achieve convergence, the network can learn discriminative feature representations with two vital characteristics, inter-class discrepancy, and intra-class cross-modality compactness.

## C. Two-Stream Local Feature Network

A typical approach [26] [30] in RGB-based person Re-ID is partitioning pedestrians into horizontal stripes to extract local feature which is concatenated to represent the body structure. Since the body structure is an intrinsic property of pedestrian, its representation is invariant for modality changes. Thus, the information about body structure is modality-shared, which can be used to learn the modality-invariant feature representation.

To this end, we propose the Two-Stream Local Feature Network (TSLFN), whose architecture is shown in Figure 4. The network contains two parts, feature extractor and feature embedding.

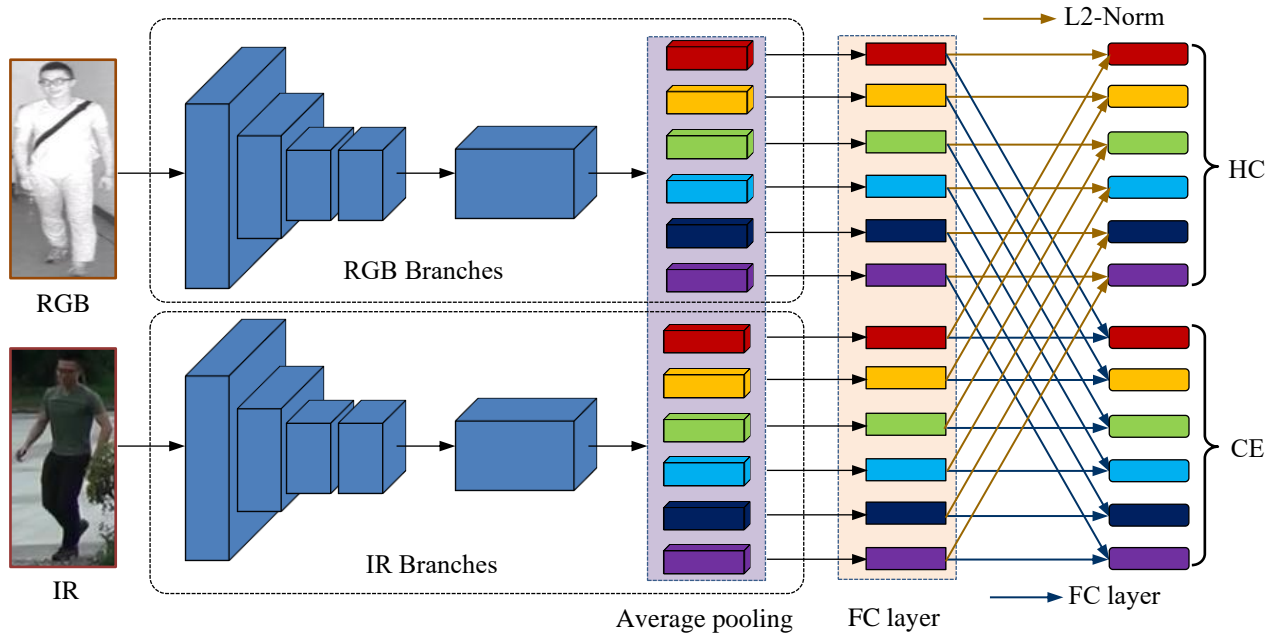

Figure 4. The architecture of the proposed Two-Stream Local Feature Network (TSLFN) with the supervision of CE loss and HC loss. The network contains two branches for two modalities. In each branch, the input images go forward the Resnet-50 backbone. Then, the feature map outputted from the backbone is split into $p$ stripes by a conventional average pooling layer. For each stripe, a weight-sharing FC layer reduces the dimension of features. Afterward, the dimension-reduced features are input into L2-Norm layers and FC layers to compute HC loss and CE loss. In the testing phase, all the dimension-reduced features are concatenated to form the final descriptor.

**Feature Extractor.** Feature extractor captures information from heterogenous images to form the final feature descriptor. As the inputs of the network include RGB images and infrared images, we adopt two individual branches to extract the information from the two modalities. With consideration of limited data, each branch contains a pre-trained backbone which inherits the architecture of Resnet50 [31] before the global average pooling layer with a slight change. The difference is that we remove the last down-sampling operation in Resnet50, which can enlarge the areas of reception fields to enrich the granularity of feature. This method has been successfully implemented in [32] [33]. Then, the feature map outputted from the backbone is uniformly partitioned into $p$ stripes in the horizontal orientation. Each stripe is averaged into a local feature vector. Afterward, we adopt a fully connected (FC) layer to reduce the dimension of each local feature vector. To bridge the gap between two modalities, corresponding fully connected layer in two branches shares the same weights. For each FC layer, we adopt a Leaky ReLU activation layer and a batch normalization layer [34] to solve the internal covariate shift problem. In the testing stage, the images are input into corresponding branches according to the modality. Then, each local feature vector undergoes L2 normalization. At last, all the feature vectors are concatenated to form the final feature descriptors. In the process of testing, given a probe image, we extract the feature descriptor of the probe and all the heterogenous gallery images. Because the identities of training images and testing images (consisting of gallery and probe images) do not overlap, we can not directly predict the labels of those gallery images. The next step is that we rank the gallery images according to the Euclidean distance of feature descriptors between the probe and all the gallery images. In the ideal condition, the heterogenous intra-class gallery images have the highest similarity. In the next section, we will use two indicators to quantitatively evaluate the performance of models.

**Feature Embedding**. The aim of feature embedding is to supervise the network learning the feature representations, which achieves the two objectives: enlarging the inter-class discrepancy and improving the intra-class cross-modality similarity as much as possible. So, we adopt HC loss and CE loss to supervise the training of the network. For HC loss, feature vectors go through L2 normalization before computing the loss value as Equation (2). With regard to CE loss, a local feature vector is inputted into a classifier, which is composed of a FC layer and a softmax activation layer. There are $p$ local feature vectors that need to be inputted to different classifiers with independent parameters. Then, the classifiers predict the identity of each feature vector, individually. For each branch, we compute CE loss according to the predicted value by the classifier and the identity of the input image. The loss for each branch is used to update the parameters of the corresponding branch in the training stage.

## IV. EXPERIMENTS

### A. Dataset description

As the first large-scale dataset for cross-modality person Re-ID, SYSU-MM01 is adopted to evaluate the effectiveness of our methods. It contains 287,628 RGB images and 15,792 IR images which are captured by 6 cameras, including four RGB cameras (Cam 1, 2, 4, 5) and two infrared cameras (Cam 3, 6). The former group works in light scenarios (day time) while the latter works when the environment is dark (night time). Except for Cam 2 and Cam 3, all the cameras are placed in different locations which can be divided into indoor and outdoor scenes. Cam2 and Cam3 are placed in the same indoor scenes. The dataset contains 491 available identities, each identity is observed by at least one RGB camera and one infrared camera. Due to the great variation among heterogenous modalities, environments, human pose, and camera viewpoint changes, the dataset is very challenging. Some examples from SYSU-MM01 are shown in Figure 1.

### B. Evaluation protocol

The experiments adopt the evaluation protocol in [8][11]. The training set consists of 22258 RGB images and 11909 infrared images of 395 persons. The testing set contains the RGB and infrared images of 96 identities. During the testing phase, RGB images in the testing set are for gallery set while infrared images are for the probe set. We adopt two testing mode to fully evaluate our methods. The first mode is all-search mode, for which all the cameras are used in the testing stage. The second mode is indoor-search, for which the cameras placed in the indoor environment are used to build the gallery set. Obviously, all-search mode is more difficult than indoor-search mode, due to the scene diversity. However indoor-search mode can evaluate the performance of cross-modality retrieval better, and the mode is more similar to the ideal condition without the drastic disturbance of environments. Therefore, the two modes are used for evaluation.

For each mode, there are two settings to form the gallery set, single-shot setting, and multi-shot setting. The difference between the two settings is the image quantity of each identity in the gallery set. One image of each identity is randomly selected to constitute the gallery set in the single-shot setting, while in the multi-shot setting, each identity contains ten images in the gallery set. Since Cam 2 and Cam 3 are placed in the same scenes, probe images captured by Cam 3 ignore the gallery images of Cam 2 in the testing phase. For each image in the probe set, we compute the feature similarity between the infrared image and every RGB image in the gallery set to match the pedestrian. We use the Euclidean distances to measure their similarity. Ideally, images of the same identity have the highest similarity. We introduce Cumulative Matching Characteristic curve (CMC) and mean Average Precision (mAP) to quantitatively evaluate the methods. Each experiment is repeated ten times with the random testing set to get average performance.

### C. Implementation details

The experiments are deployed on an NVIDIA GeForce 1080Ti GPU with Pytorch. The pedestrian images are resized to 288×144. Random cropping and random horizontal flip are used for data augmentation. The batch size is 64. To realize our proposed sampling strategy, the quantity of identity in a batch is set to 4. So, in a batch, each identity contains 8 RGB images and 8 infrared images. The output feature map of the backbone is equally split into $p=6$ stripes. The dimension of feature is reduced to 512 by the first FC layer. Thus, the dimension of the final descriptor is 6×512=3072. To balance the two loss functions, $\lambda$ is set to 0.5. SGD with momentum is adopted for optimization, in which the momentum is set to 0.9. We use decayed learning rate schedule. The learning rate is set to $1 \times 10^{-2}$ in the first 30 epochs and is decayed to $1 \times 10^{-4}$ after the 30$^{th}$ epoch.

### D. Comparison with state-of-the-art methods

We compare the proposed methods with traditional handcrafted feature based methods and deep learning based methods. The handcrafted feature based method includes HoG [35] and LOMO [3] features with different metrics: KISSME [2], LFDA [36], CCA [37], CDFE [38], GMA [41]. And the deep learning based methods are GSM [39], Zero-padding [8], TONE+HCML [9], BCTR/BDTR [11], cmGAN [10], eBDTR [44], D$^2$RL [45], DPMBN [46],

HPILN [47]. For the comparative methods, we directly copy the results from the original papers and '-' denotes the corresponding results are not reported in the original paper. The backbone used in those methods has been written in brackets. What should be mentioned is that we compare our methods with BDTR on ResNet-50 reported in [44].

The comparative results on Rank-1, 10, 20 accuracy of CMC and mAP are shown in Table 1. The results of six rows on the bottom show the performance of the proposed methods. "TSLFN(w s)+HC" refers to the Two-Stream Local Feature Network with the joint supervision of Cross-Entropy loss and Hetero-Center loss, which is the full version of the proposed methods. The denotations of the other five rows are explained in the next subsection. From Table 1, we clearly observe the superior performance of the proposed method, which greatly outperforms the existing methods in all modes. Specifically, in the most difficult mode, all-search *single-shot* mode, the performance of our method exceeds the state-of-the-art methods in term of Rank1, 10, 20 and mAP by 29.99%, 23.99%, 16.26%, and 27.15%, respectively.

TABLE I. COMPARISON WITH STATE-OF-THE-ART WORKS ON SYSU-MM01 DATASETS.

| Method | All-search | | | | | | | | Indoor-search | | | | | | | |
|---|---|---|---|---|---|---|---|---|---|---|---|---|---|---|---|---|
| | Single-shot | | | | Multi-shot | | | | Single-shot | | | | Multi-shot | | | |
| | R1 | R10 | R20 | mAP | R1 | R10 | R20 | mAP | R1 | R10 | R20 | mAP | R1 | R10 | R20 | mAP |
| HoG+Euclidean | 2.76 | 18.25 | 31.91 | 4.24 | 3.82 | 22.77 | 37.63 | 2.16 | 3.22 | 24.68 | 44.52 | 7.25 | 4.75 | 29.06 | 49.38 | 3.51 |
| HoG+KISSME | 2.12 | 16.21 | 29.13 | 3.53 | 2.79 | 18.23 | 31.25 | 1.96 | 3.11 | 25.47 | 46.47 | 7.43 | 4.10 | 29.32 | 50.59 | 3.61 |
| HoG+LFDA | 2.33 | 18.58 | 33.38 | 4.35 | 3.82 | 20.48 | 35.84 | 2.20 | 2.44 | 24.13 | 45.50 | 6.87 | 3.42 | 25.27 | 45.11 | 3.19 |
| LOMO+CCA | 2.42 | 18.22 | 32.45 | 4.19 | 2.63 | 19.68 | 34.82 | 2.15 | 4.11 | 30.60 | 52.54 | 8.83 | 4.86 | 34.40 | 57.30 | 4.47 |
| LOMO+CDFE | 3.64 | 23.18 | 37.28 | 4.53 | 4.70 | 28.22 | 43.05 | 2.28 | 5.75 | 34.35 | 54.90 | 10.19 | 7.36 | 40.38 | 60.33 | 5.64 |
| LOMO+GMA | 1.04 | 10.45 | 20.81 | 2.54 | 0.99 | 10.50 | 21.06 | 1.47 | 1.79 | 17.90 | 36.01 | 5.63 | 1.71 | 18.11 | 36.17 | 2.88 |
| GSM | 5.29 | 33.71 | 52.95 | 8.00 | 6.19 | 37.15 | 55.66 | 4.38 | 9.46 | 48.98 | 72.06 | 15.57 | 11.36 | 51.34 | 73.41 | 9.03 |
| Asymmetric FC | 9.30 | 43.26 | 60.38 | 10.82 | 13.06 | 52.11 | 69.52 | 6.68 | 14.59 | 57.94 | 78.68 | 20.33 | 20.09 | 69.37 | 85.80 | 13.04 |
| Two-stream | 11.65 | 47.99 | 65.50 | 12.85 | 16.33 | 58.35 | 74.46 | 8.03 | 15.60 | 61.18 | 81.02 | 21.49 | 22.49 | 72.22 | 88.61 | 13.92 |
| One-stream | 12.04 | 49.68 | 66.74 | 13.67 | 16.26 | 58.14 | 75.05 | 8.59 | 16.94 | 63.55 | 82.10 | 22.95 | 22.62 | 71.74 | 87.82 | 15.04 |
| Zero-padding | 14.80 | 54.12 | 71.33 | 15.95 | 19.13 | 61.40 | 78.41 | 10.89 | 20.58 | 68.38 | 85.79 | 26.92 | 24.43 | 75.86 | 91.32 | 18.64 |
| TONE+HCML | 14.32 | 53.16 | 69.17 | 16.16 | - | - | - | - | - | - | - | - | - | - | - | - |
| BCTR(AlexNet) | 16.12 | 54.90 | 71.47 | 19.15 | - | - | - | - | - | - | - | - | - | - | - | - |
| cmGAN(ResNet50) | 26.97 | 67.51 | 80.56 | 27.80 | 31.49 | 72.74 | 85.01 | 22.27 | 31.63 | 77.23 | 89.18 | 42.19 | 37.00 | 80.94 | 92.11 | 32.76 |
| BDTR(ResNet50) | 27.32 | 66.96 | 81.07 | 27.32 | - | - | - | - | 31.92 | 77.18 | 89.28 | 41.86 | - | - | - | - |
| eBDTR(ResNet50) | 27.82 | 67.34 | 81.34 | 28.42 | - | - | - | - | 32.46 | 77.42 | 89.62 | 42.46 | - | - | - | - |
| D$^2$RL(ResNet50) | 28.9 | 70.6 | 82.4 | 29.2 | - | - | - | - | - | - | - | - | - | - | - | - |
| DPMBN(ResNet50) | 37.02 | 79.46 | 89.87 | 40.28 | - | - | - | - | 44.47 | 87.12 | 95.24 | 54.51 | - | - | - | - |
| HPILN(ResNet50) | 41.36 | 84.78 | 94.51 | 42.95 | 47.56 | 88.13 | 95.98 | 36.08 | 45.77 | 91.82 | 98.46 | 56.52 | 53.05 | 93.71 | 98.93 | 47.48 |
| Baseline(w/o S) | 24.34 | 68.37 | 82.59 | 26.67 | 28.18 | 72.72 | 85.97 | 20.19 | 25.48 | 76.64 | 90.95 | 37.30 | 28.01 | 80.97 | 92.60 | 26.86 |
| Baseline(w S) | 28.52 | 72.39 | 85.26 | 30.37 | 34.18 | 77.02 | 88.47 | 23.51 | 27.79 | 76.92 | 90.53 | 38.89 | 32.42 | 83.82 | 94.66 | 28.58 |
| Baseline(w s)+HC | 41.06 | 84.40 | 93.90 | 41.88 | 46.01 | 88.17 | 95.60 | 33.99 | 44.04 | 90.74 | 97.58 | 54.29 | 53.36 | 94.18 | 98.85 | 44.90 |
| TSLFN(w/o S) | 37.20 | 81.99 | 91.50 | 38.81 | 40.74 | 85.30 | 93.93 | 31.86 | 39.48 | 85.44 | 94.20 | 49.79 | 45.52 | 90.24 | 96.98 | 40.06 |
| TSLFN(w S) | 46.78 | 86.13 | 93.18 | 46.13 | 53.18 | 90.30 | 95.84 | 39.10 | 47.39 | 87.09 | 94.08 | 55.76 | 57.14 | 93.03 | 97.70 | 46.91 |
| **TSLFN(w s)+HC** | **56.96** | **91.50** | **96.82** | **54.95** | **62.09** | **93.74** | **97.85** | **48.02** | **59.74** | **92.07** | **96.22** | **64.91** | **69.76** | **95.85** | **98.90** | **57.81** |

*E. Ablation experiments*

Our method consists of two parts, HC loss, and TSLFN. In addition, the proposed sampling strategy can improve the performance of the model supervised by CE loss. To prove the effectiveness of each component, we conduct several ablation experiments. In each experiment, unrelated settings are consistent. The results are shown in Table 1.

"baseline (w/o S)" refers to TSLFN with $p$=1, meaning that it does not partition the feature map outputted by the backbone. So, the baseline network extracts global features from input images, instead of local features. What' more, the model is only supervised by CE loss function and we adopt the sampling strategy proposed in [11], instead of our proposed sampling strategy. "baseline (w S)" refers to the baseline network with CE loss, in which we adopt the sampling strategy in the training stage. "baseline(w s)+HC" refers to the baseline network with the joint supervision of HC loss and CE loss, in which our sampling strategy is used to train the model. "TSLFN (w/o S)" refers to the Two-Stream Local Feature Network with $p$=6, in which the proposed sampling strategy and HC loss are not used in the training stage. "TSLFN (w S)" refers to TSLFN with the sampling strategy in the training phase, but we do not use the supervision of HC loss to train the model.

The comparative results between baseline and TSLFN show the performance of TSLFN outperform the baseline,

which indicates TSLFN is effective for cross-modality person Re-ID. Moreover, the comparative results between baseline and baseline+HC loss, TSLFN and TSLFN+HC demonstrate that HC loss is conducive to address the task. The results prove that HC loss can supervise the network to extract modality-shared information and improve the intra-class cross-modality similarity. Notice that the value of $\lambda$ in baseline+HC is set to 1 while $\lambda$ in TSLFN+HC is set to 0.5, because a local feature vector contains less modality-shared information than a global feature vector. In this case, the value of $\lambda$ achieving the optimal performance of baseline model is improper for TSLFN. The comparative results between baseline(w/o s) and baseline (w s), TSLFN(w/o s) and TSLFN(w s) demonstrate the sampling strategy is not only applicable for HC loss but also for CE loss.

## V. DISCUSSION

### A. Impact of $\lambda$

In this section, we conduct several experiments to investigate the influence of $\lambda$, which controls the weight of HC loss in the overall loss function. In those experiments, we vary $\lambda$ from 0.1 to 0.6, using 0.1 as the interval. The performances of different $\lambda$ on SYSU-MM01 with all-search *single-shot* mode are shown in Figure 5(a). We can observe that about 0.5 is the optimal value of $\lambda$. Besides, when the value of $\lambda$ is greater than 0.6, we find that the performance of model drops sharply. We speculate that the content of modality-shared information in local feature is not enough so that the network can not pull two modality centers closer to optimize HC loss in a correct direction, which may result in overfitting.

To verify that local features lead to the decline of the performance when $\lambda$ is set to a large value, we inquire about the impact of $\lambda$ on the baseline network as controls. The baseline network learns global feature representations, instead of local features. In the experiments on the control group, the unrelated settings are kept consistent with the experimental group and we vary $\lambda$ from 0.1 to 1, using 0.1 as the interval. The results are shown in Figure 5(b), and we can observe the phenomenon that the performance of the baseline model is improved with the increase of $\lambda$, which is different from TSLFN. Since the baseline model extracts global features and its performance is not declined with the increase of $\lambda$, the phenomenon proves our inference.

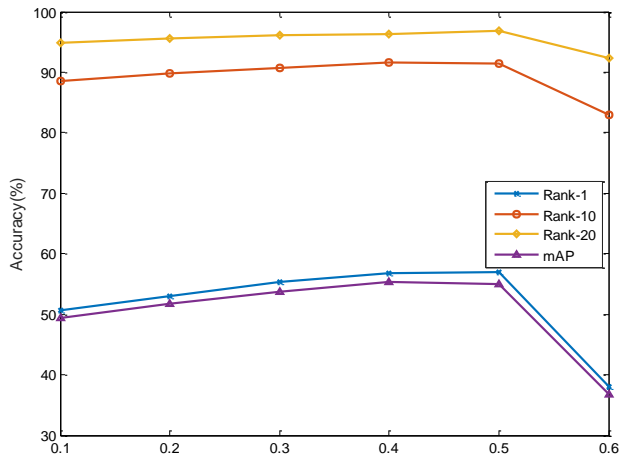

(a). Impact of $\lambda$ on TSLFN

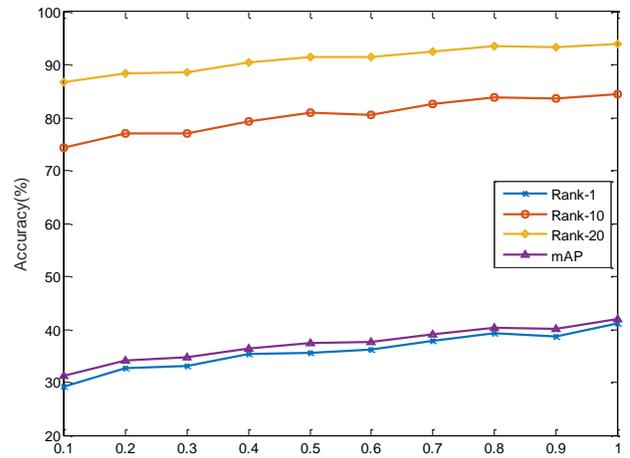

(b). Impact of $\lambda$ on baseline

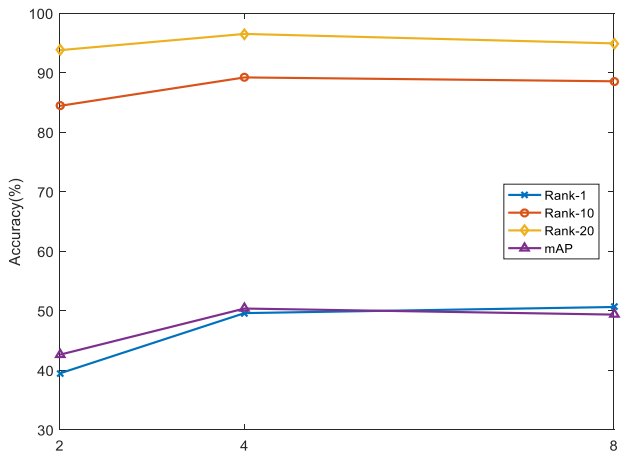

(c). Impact of $T$

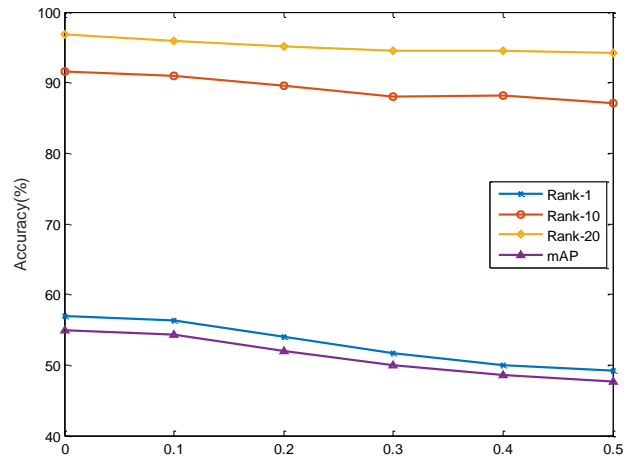

(d). Impact of $\alpha$

Figure 5. Parameter analysis. (a) the performance of TSLFN with different weights $\lambda$ of HC loss in the overall loss function. (b) the impact of $\lambda$ on baseline, in which the value range of $\lambda$ is different from (a). (c) the performance of TSLFN with different number of *T*. (d) the performance trends of TSLFN with different values of $\alpha$.

## B. The number of T

In this subsection, we conduct several experiments to investigate the influence of *T* in the sampling strategy for the performance of model. In those experiments, the batch size is fixed to 64. So, the number of RGB images and infrared images in each batch are 32. To ensure the sample quantity of each class is equal in a batch, the value of *T* is only set to 2, 4, 8, since the sample quantity of some class is less than 16. Because when $\lambda$ is set to 0.5, TSLFN with *T*=2, 4 can not achieve convergence, we set $\lambda$=0.1 for those experiments. The reason for the non-convergence may be that the computed center with too few samples can not truly reflect the center of the modality when the number of sampled images *T* is set to a small value. In this case, setting a too big value of $\lambda$ may be inappropriate. The experimental results are shown in Figure 5(c). However, what should be mentioned is that, when $\lambda$=0.1, the model with *T*=8 can not get the best performance according to Figure 5(a). The setting of the best performance of the model is $\lambda$=0.5, *T*=8. In this situation, we can observe that the performance is improved as the increase of *T*, because the centers of modality in each class can be computed correctly when *T* is set to a large value.

## C. The risk of overfitting

To investigate whether HC loss brings the extra risk of overfitting. we add a relaxation term in the definition of HC loss. The formula of modified HC loss is as

$$L_{HC} = \sum_{i=1}^{U} \left[ \|c_{i,1} - c_{i,2}\|_2^2 - \alpha \right]_+, \quad (5)$$

where $\alpha$ denotes the margin of HC loss, $[x]_+$ denotes $\max(x,0)$. When the distance between two modality centers is less than $\alpha$, the value of HC loss is zero. Thus, with the increase of $\alpha$, HC loss is easier to achieve the minimum value, the risk of overfitting is lower. The original HC loss can be regarded as the modified HC loss with $\alpha = 0$. So, we vary $\alpha$ from 0.1 to 1, using 0.1 as interval, to evaluate the risk of overfitting. The experimental results are shown in Figure 5(d), and we observe the performance of model is decreased with the improvement of $\alpha$, which indicates adopting the supervision of HC loss does not lead to overfitting of the model.

## D. Comparison among the distance metric

In HC loss, we measure the distance between two modality centers by Euclidean metric. So, the proposed formula of HC loss can be defined as

$$L_{HC} = \frac{1}{2} \sum_{i=1}^{K} D(c_{i,1}, c_{i,2}), \quad (6)$$

where $D(c_{i,1}, c_{i,2}) = \|c_{i,1} - c_{i,2}\|_2^2$. To investigate the impact of distance metrics, we use cosine similarity instead of Euclidean metric in HC loss. The following equation is the definition of cosine similarity used in HC loss,

$$D(c_{i,1}, c_{i,2}) = 1 - \frac{c_{i,1} \cdot c_{i,2}}{\|c_{i,1}\| \cdot \|c_{i,2}\|}. \quad (7)$$

We compare the performance of HC loss between Euclidean metric and cosine similarity on all-search *single-shot* mode, and the results are shown in Figure 6(a). The comparative results indicate that Euclidean metric is more suitable than cosine similarity for HC loss, partly because cosine similarity only constrains the direction of two center vectors while Euclidean metric constrains the distance between two centers.

## E. Comparison between strong and weak constraints

In HC loss, we constrain the center distance between two modality feature distributions. In the subsection, we constrain both the variance and the center to further reduce the difference between two modality distributions, which is known as the strong constraint. On the contrary, the weak constraint denotes HC loss. The strong constraint is defined as

$$L_{HC} = \sum_{i=1}^{K} \left[ \|c_{i,1} - c_{i,2}\|_2^2 + \|v_{i,1} - v_{i,2}\|_2^2 - \alpha \right]_+, \quad (8)$$

where $v_{i,1}$ and $v_{i,2}$ are the variances of two modality distributions in a mini-batch.

We conduct experiments to compare the performance of two constraints. In these experiments, other unrelated settings are consistent. We report the comparative results between strong and weak constraint on all-search *single-shot* mode in Figure 6(b). From the comparative results, we observe that the performance of strong constraints is slightly lower than the performance of weak constraints. Moreover, the computational cost for strong constraints is more expensive than it for weak constraints.

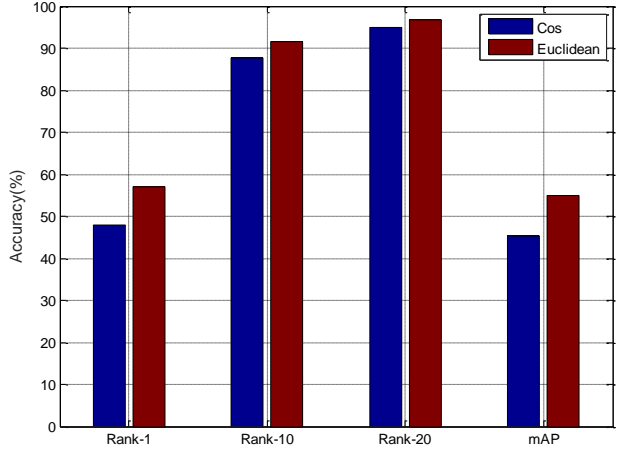
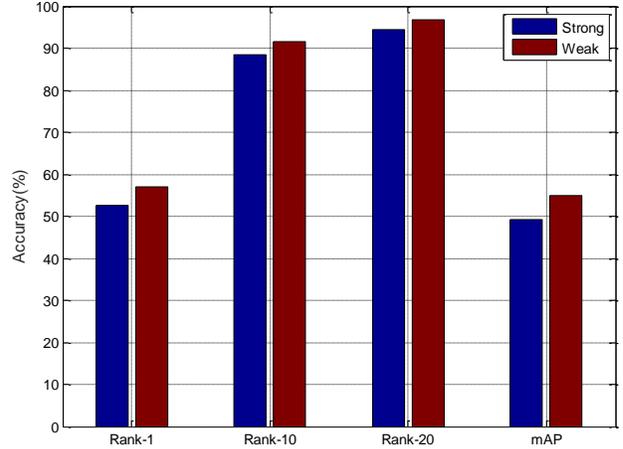

(a). Comparison of the distance metrics

(b). Comparison between strong and weak constraints

Figure 6. The comparative results with different distance metrics and constraints. (a) shows the comparative results between cosine similarity and Euclidean metric, (b) demonstrates the performance difference between strong constraints and weak constraints.

## F. The number of parts

In the subsection, we evaluate the number of parts $p$ which determines the granularity of local feature. When $p=1$, TSLFN degenerates into baseline network which extracts global feature from input images. whose performance is reported in Table 1. To reduce the influence of irrelevant variable, we adopt CE loss to train the models with different $p$, as the optimal $\lambda$ is different for different $p$. The experimental results are shown in Figure 7.

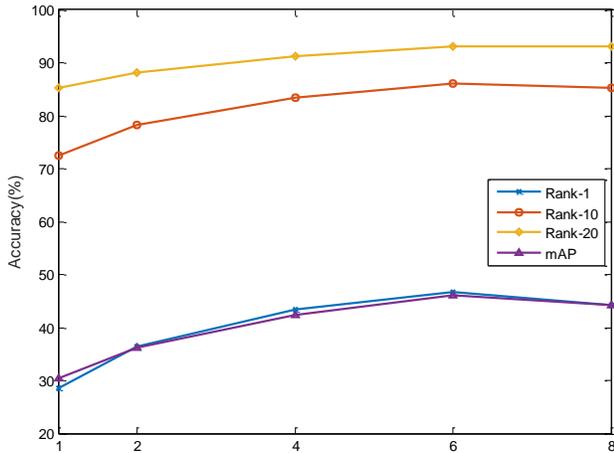

Figure 7. The impact of $p$. We demonstrate the performance of TSLFN with different $p$. When $p=1$, TSLFN is degraded to the baseline model.

We can observe that the performance of the network improves as $p$ increases at first, because the narrower granularity of local features leads to the fact that network pays more attention to the detail. However, the performance drops when $p$ is greater than 6, partly because the network can not extract efficient information with such small granularity to form a discriminative feature descriptor.

## G. The effectiveness of the proposed sampling strategy

The sampling strategy is proposed to realize HC loss, and we find that it can effectively avoid overfitting and improve the performance, especially for the model learning local feature representation. To investigate the impact of the sampling strategy for local feature learning, we use the models adopting our sampling strategy as the experimental group and the model using the sampling strategy proposed in [11] as the control group. In those experiments, we vary the number of parts $p$, using 2 as the interval.

To demonstrate the effectiveness of the sampling strategy, we only use CE loss function to train the models. The comparative results are shown in Figure 8. From the experimental results, we observe two phenomenon. First, the proposed sampling strategy is more effective comparing with the original sampling strategy [11]. Second, the benefit of the sampling strategy is enhanced with the increase of $p$.

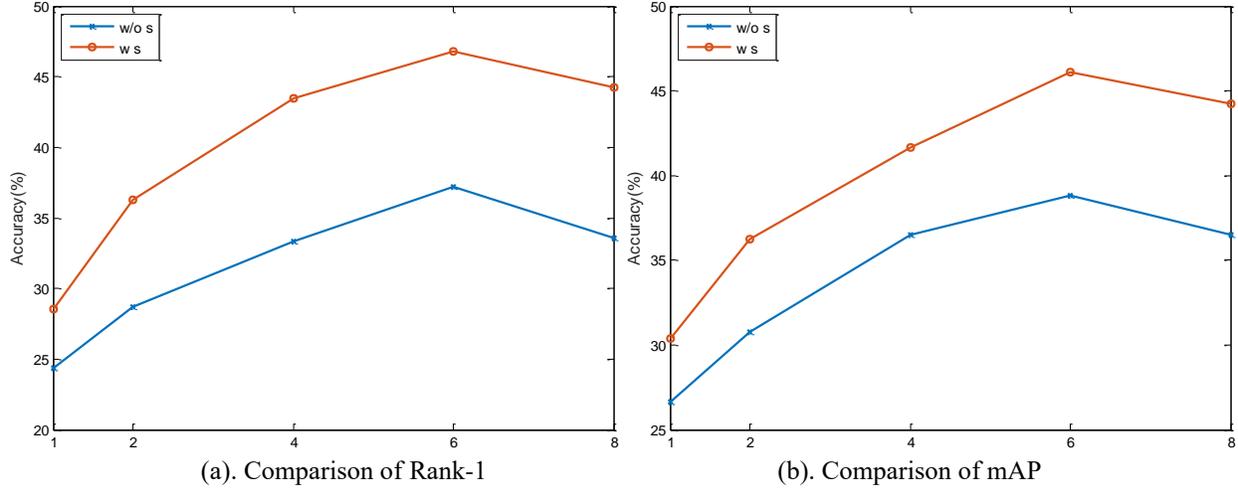

Figure 8. The impact of the sampling strategy. We compare the performance of TSLFN with our sampling strategy and the original sampling strategy in [11]. (a) and (b) are the comparative results on different indicators, in those figures, "w/o s" denotes the original sampling strategy, "w s" denotes our sampling strategy.

*H. The Comparison between HC loss and Center loss*

HC loss and center loss [29] are different, in the aspect of their aims and realization. Center loss penalizes the distance between the intra-class sample and its corresponding intra-class center to make the intra-class sample compact. However, HC loss constrains the center distance between two modality distributions, which can pull the two modality distributions close to reduce cross-modality difference. For cross-modality person re-identification, improving cross-modality similarity is more important than reducing intra-class discrepancy, because the aim of cross-modality person re-identification is that, given a query image, the trained model can retrieval the heterogenous gallery images of the same identity according to the feature similarity between the query and each gallery images. In this case, HC loss is more pertinent than center loss to the problem.

To intuitively show the difference, we illustrate the feature distribution supervised by center loss and HC loss in Figure 9. In the Figure, we observe that cross-modality difference of feature with center loss is bigger than that with HC loss, which is reflected by the comparison of the center distance between the two feature distributions.

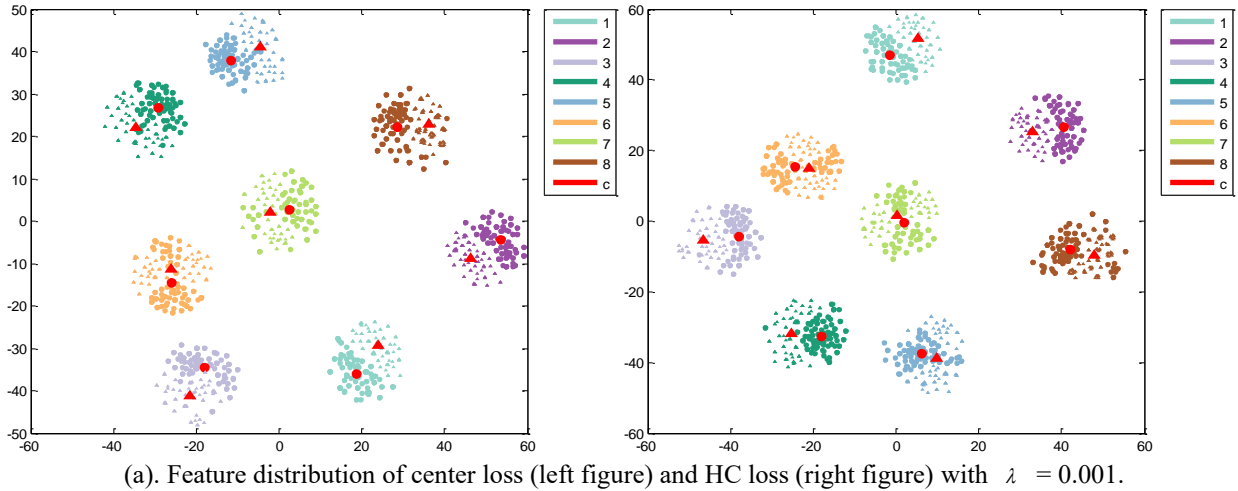

(a). Feature distribution of center loss (left figure) and HC loss (right figure) with $\lambda = 0.001$.

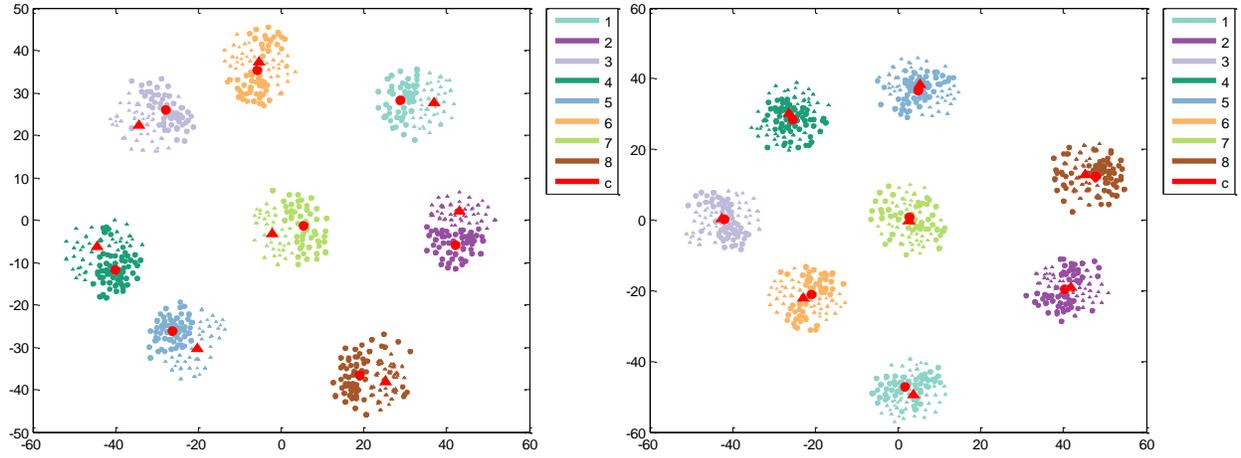
(b). Feature distribution of center loss (left figure) and HC loss (right figure) with $\lambda = 0.01$.

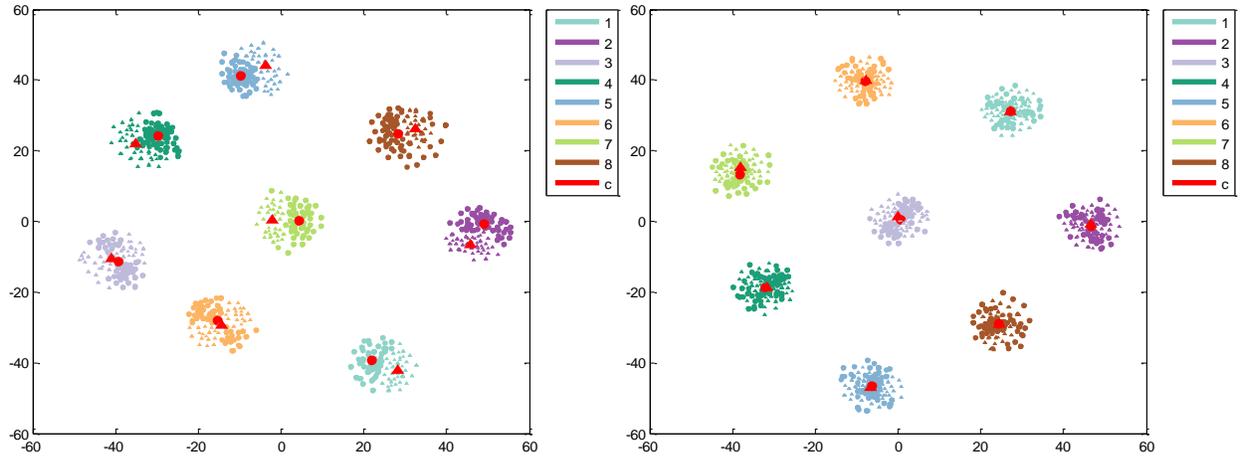
(c). Feature distribution of center loss (left figure) and HC loss (right figure) with $\lambda = 0.1$.

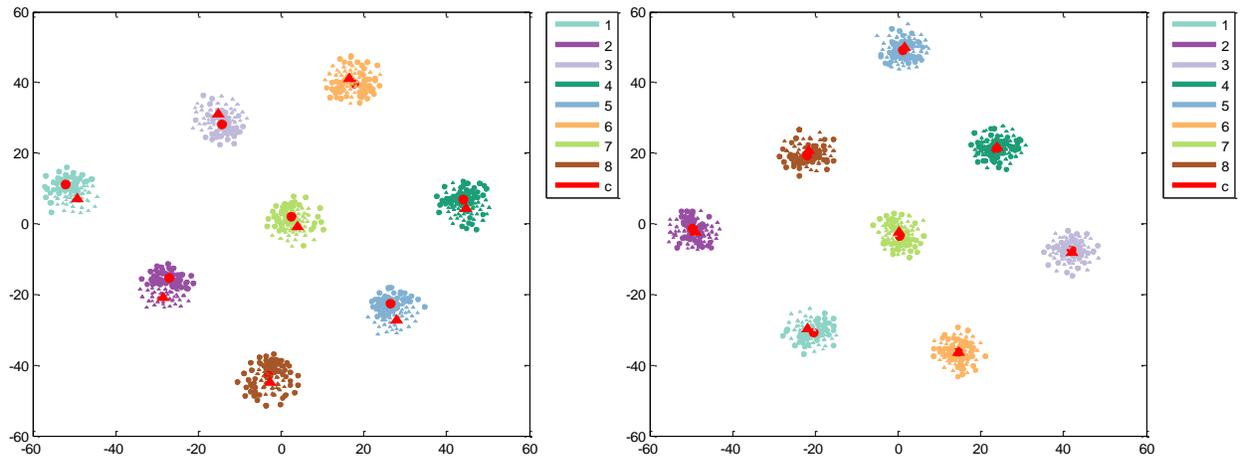
(d). Feature distribution of center loss (left figure) and HC loss (right figure) with $\lambda = 1$.

Figure 9. Feature distributions between center loss (left figure) and HC loss (right figure). We illustrate the changes of feature distribution with the increase of $\lambda$, from 0.001 to 1. In those figures, we observe that the center distance between two modalities with the supervision of center loss is bigger than the supervision of HC loss with the same $\lambda$.

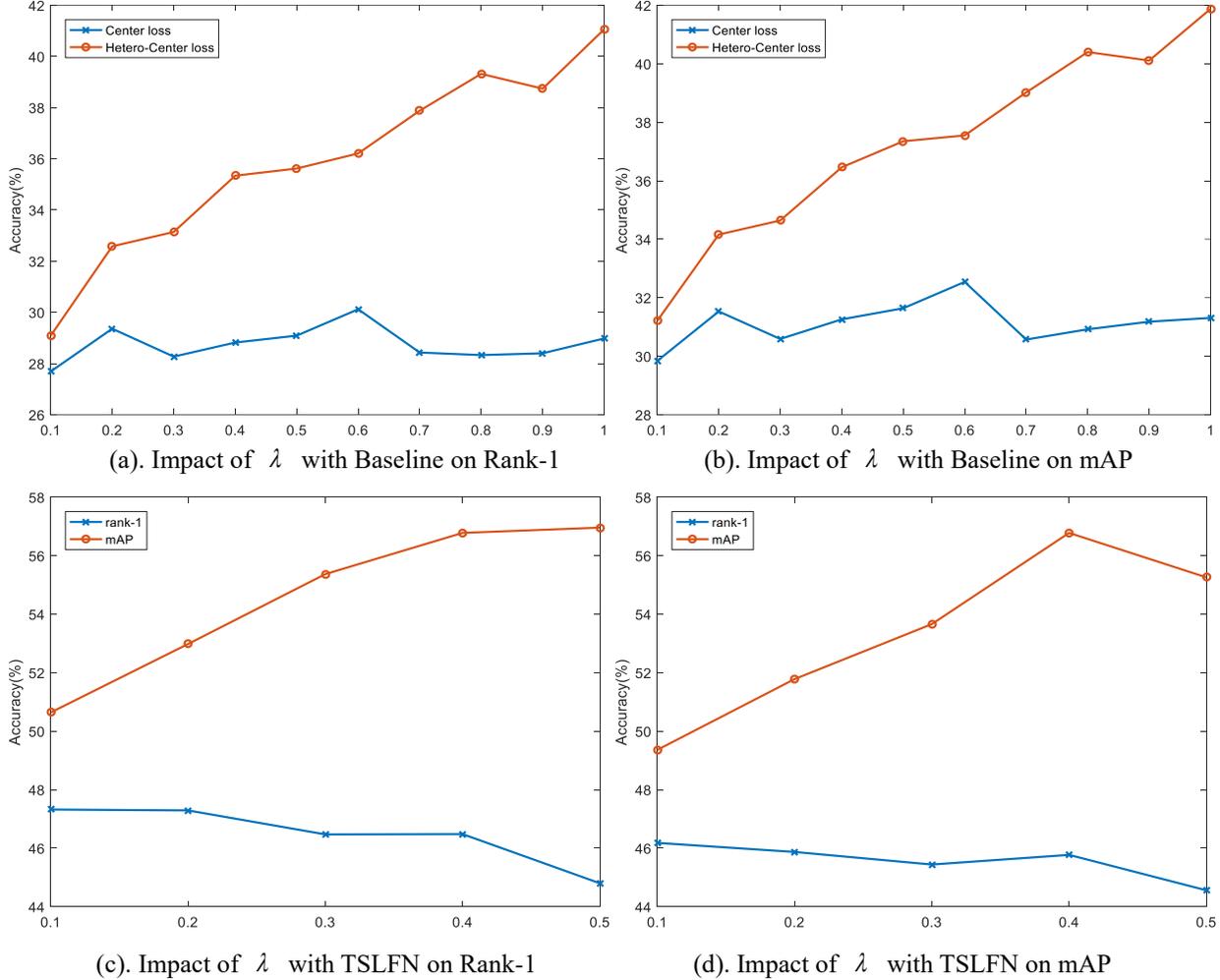

Figure 10. The comparison between HC loss and center loss on baseline and TSLFN. (a) and (b) are the comparative results of different indicators (rank-1 and mAP) on baseline, (c) and (d) are the comparative results of different indicators on TSLFN. In the figures, we observe that HC loss outperform center loss by a large margin and the margin is improved with the increase of $\lambda$.

What's more, to prove that HC loss is more suited to the task than center loss, we conduct the experiments to directly compare the performance between the two methods. The experiments adopt SYSU-MM01 dataset with the difficult mode, all-search *single-shot* mode. In the experiments, the learning rate $\alpha$ of center in center loss is set to 0.3, because we find that too big value of $\alpha$ cause non-convergence of models in the training phase. What's more, the increase of $\alpha$ does not bring the improvement of model performance, which is also observed in [29]. For fairness and comprehensiveness, we compare the performance of the two methods on different network structures, baseline and TSLFN. And, we also compare the performance of the two loss functions with different $\lambda$. We vary $\lambda$ from 0.1 to 1 in the experiments of baseline, and vary $\lambda$ from 0.1 to 0.5 in the experiments on TSLFN. The experimental results are shown in Figure 10, which demonstrates the performance of our methods exceeds center loss by a large margin.

## VI. CONCLUSIONS

In this work, we propose a novel loss function called Hetero-Center (HC) loss for cross-modality person Re-ID task. With the joint supervision of CE loss and HC loss, the model directly learns feature representations achieving the vital aim, inter-class discrepancy and intra-class cross-modality similarity simultaneously. Moreover, we propose a network architecture named Two-Stream Local Feature Network (TSLFN) to learn discriminative local feature representations from heterogenous images. The framework has advanced performance and simple structure, proving itself as an excellent baseline for future work. Extensive experiments strongly demonstrate the effectiveness of the proposed methods, which greatly outperform state-of-the-art works.

ACKNOWLEDGMENTS

This research was supported by National Natural

Science Foundation of China under Grant 61501177, 61772455, 61902084, Guangzhou University's training program for excellent new-recruited doctors (No. YB201712), Guangdong Natural Science Foundation under Grant 2017A030310639, Featured Innovation Project of Guangdong Education Department under Grant 2018KTSCX174, Yunnan Natural Science Funds under Grant 2018FY001(-013), the Program for Excellent Young Talents of National Natural Science Foundation of Yunnan University(2018YDJQ004).

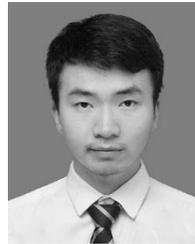

**Yuanxin Zhu** received the B. E. degree in Electronics and Information Engineering from Guangzhou University in 2019. He has worked in Laboratory of Audio-video and Lighting Technology in Guangzhou University for 2 years. His area of research includes human detection, generative adversarial network, and person re-identification.

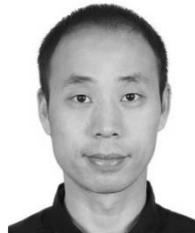

**Zhao Yang** received a Ph.D. degree from South China University of Technology in 2014. He is currently a lecturer of School of Mechanical and Electric Engineering, Guangzhou University. His research interests include machine learning, pattern recognition.

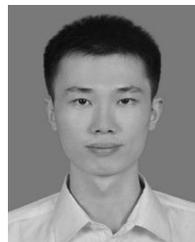

**Li Wang** received a bachelor degree in electronic engineering from Southeast University, China, in 2009 and received a Ph.D. degree in physical electronics from Southeast University, China, in 2015. He is now working in the School of Mechanical and Electric Engineering, Guangzhou University. His research interests include brain-computer interface, biomedical signal processing, pattern recognition.

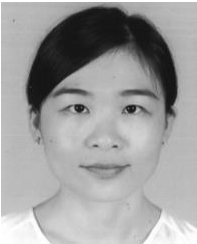

**Sai Zhao** received the Ph. D degree in communication and information system from Sun Yat-Sen University (SYSU), Guangzhou, China, in 2015, and the Master and Bachelor degrees in Communication Engineering from Central South University, Changsha, China, in 2006 and 2003, respectively. She is currently a lecturer at Guangzhou University, Guangzhou, China. Her current research interests include machine learning in wireless communication, convex optimization, physical layer security and non-orthogonal multiple access.

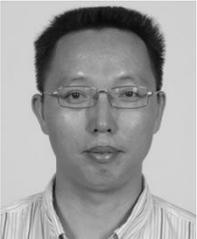

**Xiao Hu** received a M.S. degree in communication and information system from Yunnan University, Kunming, China, in 2003 and received a Ph.D. degree in Biomedical Engineering from Shanghai Jiao Tong University, Shanghai China, in 2006. He is currently a Professor with the School of Mechanical and Electric Engineering, Guangzhou University, and as a visiting scholar working at the University of Queensland from Jan. 2016 to Jan. 2017. His current research interests include machine vision, image and biomedical signal processing, pattern recognition, etc.

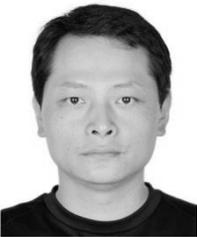

**Dapeng Tao** received a B.E degree from Northwestern Polytechnical University and a Ph.D. degree from South China University of Technology, respectively. He is currently with School of Information Science and Engineering, Yunnan University, Kunming, China, as a Professor. He has authored and co-authored more than 50 scientific articles. He has served more than 10 international journals including IEEE TNNLS, IEEE TMM, IEEE CSVT, IEEE SPL, and Information Sciences. Over the past years, his research interests include machine learning, computer vision and robotics.